\documentclass[runningheads]{llncs}
\usepackage[T1]{fontenc}

\usepackage{graphicx}

\usepackage{comment}
\usepackage{multirow}
\usepackage{hyperref}
\usepackage{tikz}
\usepackage{float}
\usepackage{caption}
\usepackage{url}
\usepackage{graphicx}
\usepackage{booktabs}
\usepackage{longtable}
\usepackage{needspace}

\usetikzlibrary{shapes,arrows.meta,positioning, fit, shapes.geometric}
\tikzset{
    block/.style={
        rectangle,
        draw,
        fill=blue!30,
        text width=2cm,
        text centered,
        minimum height=1cm,
        font=\scriptsize
    },
    line/.style={
        draw,
        -Latex
    },
    auto,
}

\begin{document}
%

\title{ETLNet: An Efficient TCN-BiLSTM Network for Road Anomaly Detection Using Smartphone Sensors}
\titlerunning{ETLNet}{}

%
%
\author{Mohd Faiz Ansari\inst{1} \and
        Rakshit Sandilya\inst{1} \and
        Mohammed Javed\inst{1} \and
        David Doermann\inst{2}}
\authorrunning{Faiz et al.} 
\institute{Computer Vision and Biometrics Lab, Department of IT, IIIT Allahabad, India \and
           Computer Science and Engineering, University at Buffalo, Buffalo, NY, USA \\
           \email{\{mml2022006, mml2022008, javed\}@iiita.ac.in}, doermann@buffalo.edu}

\maketitle              
\begin{abstract}
Road anomalies can be defined as irregularities on the road surface or in the surface itself. Some may be intentional (such as speedbumps), accidental (such as materials falling off a truck), or the result of roads' excessive use or low or no maintenance, such as potholes. Despite their varying origins, these irregularities often harm vehicles substantially. Speed bumps are intentionally placed for safety but are dangerous due to their non-standard shape, size, and lack of proper markings. Potholes are unintentional and can also cause severe damage. To address the detection of these anomalies, we need an automated road monitoring system. Today, various systems exist that use visual information to track these anomalies. Still, due to poor lighting conditions and improper or missing markings, they may go undetected and have severe consequences for public transport, automated vehicles, etc. In this paper, the Enhanced Temporal-BiLSTM Network (ETLNet) is introduced as a novel approach that integrates two Temporal Convolutional Network (TCN) layers with a Bidirectional Long Short-Term Memory (BiLSTM) layer. This combination is tailored to detect anomalies effectively irrespective of lighting conditions, as it depends not on visuals but smartphone inertial sensor data. Our methodology employs accelerometer and gyroscope sensors, typically in smartphones, to gather data on road conditions. Empirical evaluations demonstrate that the ETLNet model maintains an F1-score for detecting speed bumps of 99.3\%. The ETLNet model's robustness and efficiency significantly advance automated road surface monitoring technologies.

\keywords{road anomaly detection  \and deep learning \and smartphone sensors.}
\end{abstract}

\section{Introduction}
\label{sec:1}


Traffic accidents have become an increasingly pressing concern globally, with the deterioration of road surfaces contributing significantly to this problem. Key insights into this issue come from various sources. In 2022, over 4.4 thousand accidents were caused by
potholes\cite{1}.There were 30 crashes and nine deaths a day due to speed breakers, according to Times of India report\cite{2}.

We can detect the road anomalies in various ways, such as manual road surveys, cameras, inertial sensors, etc \cite{24}. Manual surveys could be more efficient. The choices left are either visual or inertial sensors. Both are good in their ways. Many advanced algorithms have already been developed that use cameras to detect road anomalies. The problem with visual-based detection requires good lighting conditions and accurate markings of speed bump stripes, but in India, there are two types of speed bumps, one with markings and the other without any markings\cite{20}. The visual-based detection methods occasionally missed the anomalies due to environmental factors such as unusual speed bump design, high vehicle speed, camera blind spot, sun glare or reflection, shadows \cite{21}, obstructions like leaves, snow, debris, etc. The strong side of using vision-based detection is that you know where the anomaly is in advance. The inertial sensors offer several advantages over active sensors like cameras, LiDAR, etc \cite{23}. The inertial sensors are cost-effective, robust to environmental conditions, can be used to measure vehicle response directly, have low power consumption, and have ease of data processing. The downside is it doesn't tell in advance about the anomalies.


Nowadays, almost everybody has a smartphone, and these smartphones are equipped with advanced sensors. 
People use smartphones to plan their routes, so if we somehow already make these anomalies visible on their maps and inform the drivers about upcoming anomalies, it will surely reduce the effect of these anomalies and alert the municipal corporation about the same for quick maintenance.
The manual methods to detect these anomalies are based on calculating the IRI - International Roughness Index \cite{19} obtained from measured longitudinal road profiles or visual inspection by a human. These methods could be more efficient as they require more resources, time, and human involvement. Nowadays, studies focused on solving this problem use machine-learning and deep-learning approaches as they are efficient and accurate.
Various methods are available, such as threshold-based, ML-based, DL-based, etc. In the threshold-based method, passive sensors like accelerometer data are recorded and set a threshold \cite{22}. A value above it means an anomaly\cite{3}.In ML-based methods, RF(Random Forest), RT(Regression Tree), SVM(Support Vector Machine), etc are used\cite{4}. In DL-based approach ANN\cite{5}, CNN\cite{6}, LSTM\cite{6}, GRU\cite{6}, etc are used.

The proposed TCN-BiLSTM network focuses on detecting speed bumps from the accelerometer, gyroscope, and speed readings from smartphone sensors.This\\ smartphone-based speed bump detection classifies road anomalies as speed bumps or no speed bumps. The ETLNet (Enhanced Temporal-BiLSTM Network) extracts the features with two TCN layers and 1 BiLSTM layer. The dense and sigmoid layers classify the sensor data from the extracted features after the TCN-BiLSTM network. The values added by the approach are as follows:
\begin{itemize}
    \item This paper presents a novel approach, the Enhanced Temporal-BiLSTM Network (ETLNet), which integrates two Temporal Convolutional Network (TCN) layers with a Bidirectional Long Short-Term Memory (BiLSTM) layer.
    \item This model is tailored to detect anomalies effectively irrespective of lighting conditions, as it depends not on visuals but on inertial sensor data collected from smartphones.
    \item ETLNet-based anomaly detection predicts whether the road has a speed bump using the data from inertial sensors.
    \item To measure the performance of the proposed ETLNet, we use accuracy, precision, recall, and f1-score metrics on the speed bump dataset.
\end{itemize}

The rest of the paper has sections: the related work done in this field is given in section \hyperref[sec:2]{2}. In section \hyperref[sec:3]{3}, we discuss the proposed approach and the novel architecture for speed bump detection in detail. In section \hyperref[sec:4]{4}, the experimental results are shown for the proposed model, and the dataset used is discussed there. In section \hyperref[sec:5]{5} ablation study is provided. Finally, in section \hyperref[sec:6]{6}, we summarize the conclusion of the work.

   
    
    

\section{Related Work}
\label{sec:2}
In recent years, several methods have been introduced to detect road anomalies. In a broader perspective, they are divided into two types: the first is by using visuals like images, videos, etc., and the second one is the use of passive sensors like inertial sensors, for example, accelerometers, gyroscopes, etc. Works related to the approach of sensors, specifically in smartphones, are discussed as follows: 

Two approaches are used to get the inertial sensor data: from smartphone sensors\cite{8} or the sensors mounted on the vehicle. Aljaafreh et al.\cite{7} proposed a speed bump detection model based on the Fuzzy Inference System. The features they used were accelerometer readings and the vehicle's speed from the smartphone. They only used the vertical component of acceleration. Their system calculates the signal variance, the first input to fuzzy logic, while the vehicle's speed is the second input. The result of the Fuzzy logic was the boolean speed bump detection. They tested their model under different vehicle speeds and concluded that the bump detection accuracy was high in the middle of the bump. Aragon et al.\cite{8} presented a model that could classify road anomalies, such as speed bumps, asphalt speed bumps, and potholes. The only feature they used was accelerometer readings from the smartphone. They used SVM to segment the data into anomaly and non-anomaly classes. The signal segments were classified using a Bag of Words and Random Forest representation, producing an accuracy of 75\%.

In the work of Rishiwal et al., \cite{3}, speed bumps and potholes were detected by setting some threshold values based on the vehicle's speed. The model classifies the anomalies along with the severity level. Data were collected using a smartphone. They use only accelerometer readings for classification purposes and GPS to track the anomalies. The accuracy of the proposed approach was 93.75\%. Savera et al.\cite{11} developed an Android application to detect and alert about speed bumps and ditches within a certain distance. They collect data from various devices (android smartphones) and several vehicles. They use only a single feature, i.e., accelerometer readings, for their model to detect the anomaly and GPS to get the track of coordinates of anomalies. They also take care of the orientation of smartphones. A Support Vector Machine is trained on the collected data, producing an accuracy of 85\%. Verma et al.\cite{12} proposed a novel generalized outdoor localization scheme, UrbanEye. The goal is to estimate the travel time and current location by calculating the turns, stops, and speed bumps. This study uses accelerometer, gyroscope, compass, and smartphone GPS as data sources. They used accelerometer reading (single component) and a threshold value to detect speed bumps. The accuracy of their proposed work comes out to be 87.5\%. 

Souza et al.\cite{13} developed an application to detect the road surface type, pavement conservation condition, obstacles, and speed bumps. The accelerometer readings (all three components) from the smartphone are used. The paper proposed a novel approach combining warping distance measures with Complexity Invariant Distance (CID) for better classification across various scenarios. The paper also evaluated the effectiveness of using all acceleration components separately versus using a single series representing overall acceleration magnitude. The proposed method detects the speed bumps with an accuracy of 81.3\%, precision of 87\%, a recall of 93\%, and an F1-score of 90\%. Dey et al.\cite{4} proposed a model based on machine learning to detect the road anomalies such as smooth roads, uneven roads, potholes, speed humps, and rumble strips. As for the data, they used accelerometer and magnetometer readings from the smartphone, which were placed at different locations such as the dashboard, cupboard, dock, and inside the pocket. The study was carried out using different machine learning models for classification like Random Forest (RF), Random Tree (RT), and Support Vector Machine (SVM). They used Greedy Stepwise, Ranker, and BestFirst Algorithms for the feature selection. They used minimum, maximum, median, and standard deviation statistical features. The accuracy of the proposed model in this study comes out to be 92\%.

Kumar et al.\cite{15} proposed a model that can classify the road as smooth, speed bump, or rumble strip. The model uses data from the smartphone's accelerometer and gyroscope, fixed on the vehicle's dashboard. The proposed model is based on an artificial neural network (ANN), which can detect speed bumps with an accuracy of 98\%, a precision of 98\%, recall of 98\%, and an F1-score of 98\%. Kempaiah et al.\cite{16} proposed a deep learning model that can classify between three classes, road, pothole, and speed bump, and uses a Recurrent Neural Network, i.e., LSTM (Long Short-term Memory) with ADAM optimizer. The data fed into the model was collected from the accelerometer and GPS module. The proposed approach worked in two phases: analysis and notification. The analysis phase detects the anomaly and uploads the results to the cloud, and then the notification phase notifies us of any upcoming anomaly nearby. The accuracy and precision of the model are 80\% and 81\%, respectively. Menegazzo et al.\cite{6} performed a pervasive study on speed bump detection using inertial sensors. The paper proposed various DNN (Deep neural network) based models to detect speed bumps like BiLSTM, GRU, CNN-LSTM, and ConvLSTM. The models get trained on nine datasets with different vehicles and drivers. Their best model is CNN-LSTM, with an accuracy of 98.59\% and an f1-score of 99.19\% for a speed bump.

Salman and Mian\cite{25} developed a deep learning-based system using data from smartphone sensors in a car to detect and categorize speed bumps into sinusoidal, flat-top, thump, and round-top types. They trained and fine-tuned CNN and LSTM models, finding that CNN outperformed LSTM, achieving 92\% for bumps, with overall detection and characterization accuracies of 98.92\% (combining both normal and bump) and 95\%, respectively. Testing yielded 99.6\% detection accuracy and 95.8\% characterization accuracy, with only two false positives. To enhance reliability, they implemented a suppression mechanism to minimize consecutive false detections from overlapping samples. While CNN and LSTM are standard models, their innovation lies in the detailed categorization of bump types.

\begin{figure}
    \centering
    \includegraphics[width=0.9\linewidth]{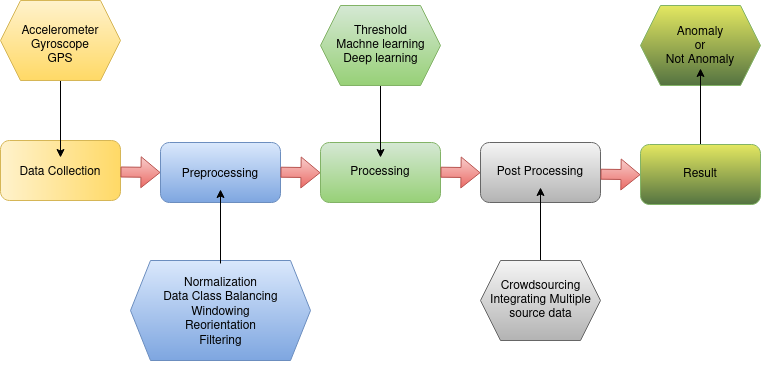}
    \caption{Flowchart illustrating the process of speed bump detection.}
    \label{fig:process}
\end{figure}

Most proposed models in the above studies need a large dataset or more generalization. Menegazzo et al.\cite{6} proposed model is well generalized but used externally fitted sensor data from multiple locations like left and right side readings for below speed bump, above speed bump, and dashboard. The proposed model(ETLNet) used data from a single location (dashboard) and one side (right) and is well generalized. Fig. \ref{fig:process} summarizes the processes from the related works.

\section{Proposed Method}
\label{sec:3}
The proposed model has two key blocks. The first block contains two sets. The pattern analysis unit is built with a Temporal Convolutional Network layer, TCN. We use a batch normalization layer for consistent data handling, and to prevent overfitting, we use a dropout layer.

The second block focuses on sequence mapping, understanding, and interpreting data sequences. This contains an BiLSTM layer, ideal for recognizing the patterns over time. It mirrors the first block's batch normalization and dropout layers, which helps maintain data integrity and model stability.

Following these blocks, the model integrates a dense layer with a ReLU activation function, aiding in nuanced decision-making. This is followed by another dense layer with a sigmoid activation, fine-tuning our output for binary classification.

\begin{figure}[H]
    \centering
    \includegraphics[width=1\linewidth]{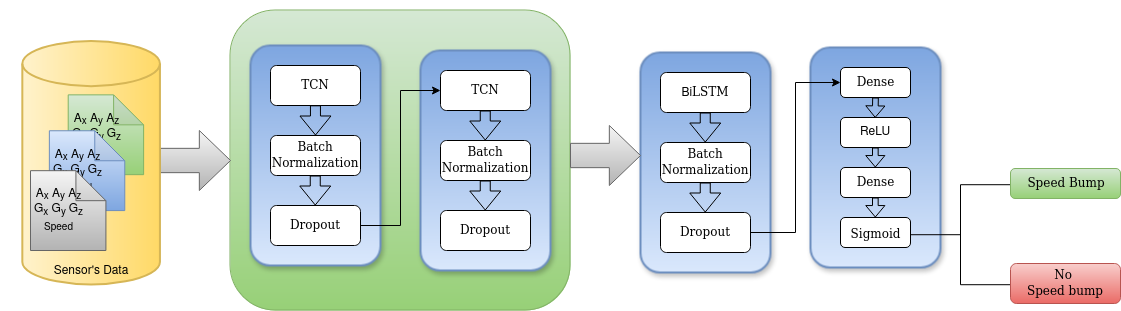}
    \caption{Proposed architecture of the ETLNet model for speed bump detection.}
    \label{fig:2}
\end{figure}

The model culminates with an output layer, where the processed data leads to our final results. This structure is crafted to balance detailed pattern analysis with effective sequence mapping. The block diagram is shown in Fig. \ref{fig:2}.

\vspace{-12pt}
The primary function of the TCN layer here is to identify speed bump patterns across time by analyzing how different data points in the input sequence related to each other over time. When we pass the input sequence through the first TCN layer, the model focuses on extracting the time-dependent features from that sequence, laying the groundwork for deeper temporal analysis. On passing the output of the first TCN layer as an input to the second TCN layer, the model further refines the extracted features, enhancing the depth and precision of the temporal patterns identified by the initial TCN layer.

Now, the sequence passes through the Bidirectional Long Short-Term Memory (BiLSTM) layer. This layer captures and retains the long-term dependencies from the output sequence of the second TCN layer. It is an essential function for understanding complex and extended patterns.

The output sequence from the BiLSTM layer passes through the first dense layer, where the data starts to be consolidated. This dense layer sets the stage for decision-making processes. Now, the sequence transforms the ReLU (Rectified Linear Unit) activation function. This step helps the model get the nuances and see the finer details in the learned patterns. The last stage of the sequence's journey involves another dense layer, followed by a sigmoid activation layer. This sigmoid function normalizes the output, making it suitable for speed bump detection tasks.

\section{Experimental Protocol}
\label{sec:4}
\subsection{Dataset}

The dataset we used is Passive Vehicular Sensors Dataset (PVS) \cite{18} is a set of nine datasets with three distinct cars, drivers, and situations.
The data preprocessing step consists of defining the input and output features, normalizing the data features, data windowing, data division for training and validation, and data class balancing \cite{6}. In this study, we use accelerometer(X, Y, Z), gyroscope(X, Y, Z), and speed data, total seven features \cite{6} from the dashboard location, only from the right side (as in India, the right side is the driving side) to simulate that the data is coming from smartphone sensors placed on the dashboard. This choice helps us realistically mimic how a smartphone, placed in a similar dashboard position, would capture data. 

\subsection{Training and Testing}
There are several ways to divide the PVS dataset for training and testing; one is to leave one out of cross-validation, and the other is to split the dataset into two distinct sets, ensuring they differ in vehicle types, drivers, and scenarios such that the testing set is unseen to model

\subsection{Evaluation Metrics}
To evaluate the performance of our proposed model ETLNet, we use validation accuracy, precision, recall, and f1-score.

Validation accuracy measures how well our model performs on a separate dataset that it hasn't seen during training. It better indicates our model's effectiveness in the real world. 

Precision tells us that all the instances in our model predicted speed bumps were correctly predicted. A higher precision means that when our model predicts a speed bump, we can be more confident that there is a speed bump.

Recall tells us of all the actual speed bumps present and how many my model successfully detects. A high recall means our model effectively detects most of the speed bumps. 

The F1-score combines precision and recall into a single number, offering a balanced measure of a model's accuracy, especially when dealing with imbalanced datasets like in our condition.
\section{Experiments}
\subsection{Comprehensive Analysis of Model Efficacy Across Varied Window Sizes and Sensor Locations}

This study delves into an in-depth examination of three distinct models - 3-BiLSTM, 3-TCN, and our proposed ETLNet model - in their roles for detecting speed bumps using sensors on various vehicle types. Here 3-BiLSTM means 3 layers of BiLSTM each followed by batch normalization and dropout and same for 3-TCN. Our systematic assessment spanned different sensor positions, meticulously measuring key performance indicators: validation accuracy, precision, recall, and F1-score. A critical aspect of our analysis focused on the impact of varying window sizes on these metrics for each respective model.

\subsubsection{Experiment 1:} Identifying Optimal Window Sizes for Enhanced Model Accuracy in Speed Bump Detection.

The optimal results from our initial experiment, as elaborated in Table \ref{tab:2}, provide a detailed perspective on how different window sizes influence each model's performance at specific sensor locations. Our findings reveal the proposed ETLNet model's consistent superiority in all measured metrics across various configurations, surpassing the BiLSTM and TCN models.

For the BiLSTM model, a window size of 300 emerged as the most effective, consistently yielding the best outcomes for all vehicle types at various sensor positions.
The TCN model, however, exhibited a sensitivity to window size adjustments: a window size of 200 was optimal for Car 1; for Car 2, the ideal size was 300; and for Car 3, a larger window size of 500 was most effective.
The ETLNet model, impressively maintaining high performance with a uniform window size of 300, underscored its adaptability and robustness across different vehicle types and sensor positions.

\subsubsection{Experiment 2:} Aggregated Sensor Position Analysis

In our second experimental stage (Table \ref{tab:3}), we calculated the average values of each metric for each sensor position across the different cars. This broader analysis further accentuated the proposed ETLNet model's overall superior performance, achieving the highest average metrics for each car with a consistent window size of 300.

\subsubsection{Experiment 3:} Evaluating Cross-Vehicle Generalization and Window Size Influence

Our third experiment, which involved training models with aggregated data from all vehicles (shown in Table \ref{tab:4}), was critical for understanding the influence of window size on each model's performance in various sensor positions. This experiment was crucial for unraveling each model's ability to adapt and optimize across a diverse range of data.

For the BiLSTM and TCN models, we found that a window size of 300 was ideal at the suspension sensor position below, achieving top results in validation accuracy, recall, and F1 score. Notably, window size 300 proved the most efficient for precision in BiLSTM and TCN models at the above suspension sensor position. This finding highlights the intricate relationship between window size and specific model performance metrics. On the other hand, for other metrics at the above suspension and dashboard positions, increasing the window size to 400 enhanced the performance of these models. Consistent with its previous robust performance, our proposed ETLNet model, employing a window size 400, demonstrated superior performance across all sensor positions and metrics. 

\subsection{Comparative Analysis of Speed Bump Detection Models}
In examining Table \ref{tab:5}, we're invited to step back and appreciate the strides made in speed bump detection over the years. The 2018 study by Souza et al. \cite{13}. provided an accuracy of 81.13\%. On the other hand, Dey et al. \cite{4}. in 2019 improved the accuracy to 92\%. In 2022, Kumar et al. \cite{15}. They achieved 98\% accuracy uniformly across all metrics presenting their ANN model. Though diverging from using smartphone sensor readings, Kempaiah et al. \cite{16}. and Menegazzo et al. \cite{6}. brought their flavors to the table. Kempaiah et al. \cite{16}.'s LSTM approach, while not reaching the heights of others, offers valuable insights. In contrast, Menegazzo et al. \cite{6}.'s CNN-LSTM model stands out for its high accuracy, impressive precision, and recall, which paints a picture of a highly reliable model. Our proposed ETLNet achieves an even higher accuracy of 98.84\%, complemented by the highest scores in precision, recall, and F1-score seen in Tables \ref{tab:2}, \ref{tab:3}, \ref{tab:4}, \ref{tab:5}.

{\small
\begin{longtable}{|c|c|c|c|c|c|c|}

\caption{Metrics Across Sensor Positions for Cars 1, 2, and 3 (3-BiLSTM, 3-TCN, ETLNet)} \label{tab:2} \\

\hline
\textbf{Car} & \textbf{Sensor Position} & \textbf{Metric} & \textbf{3-BiLSTM} & \textbf{3-TCN} & \textbf{ETLNet} \\
\hline
\endfirsthead
\multicolumn{6}{c}%
{\tablename\ \thetable\ -- \textit{Continued from previous page}} \\
\hline
\textbf{Car} & \textbf{Sensor Position} & \textbf{Metric} & \textbf{3-BiLSTM} & \textbf{3-TCN} & \textbf{ETLNet} \\
\hline
\endhead
\hline \multicolumn{6}{|r|}{\textit{Continued on next page}} \\ \hline
\endfoot
\endlastfoot
\multirow{12}{*}{Car 1} & \multirow{4}{*}{Below Suspension} & Val Accuracy  &94.50  &95.061  &\textbf{98.80}  \\ \cline{3-6}
                        &                                   & Precision &98.38  &96.60  &\textbf{98.50}  \\ \cline{3-6}
                        &                                   & Recall    &95.15  &96.80  &\textbf{99.14}  \\ \cline{3-6}
                        &                                   & F1-Score  &96.68  &96.68  &\textbf{98.82}  \\ \cline{2-6}
                        & \multirow{4}{*}{Above Suspension} & Val Accuracy  &95.24  &90.72  &\textbf{98.57}  \\ \cline{3-6}
                        &                                   & Precision &98.24  &98.86  &\textbf{99.06}  \\ \cline{3-6}
                        &                                   & Recall    &96.17  &89.05  &\textbf{99.26}  \\ \cline{3-6}
                        &                                   & F1-Score  &97.17  &93.45  &\textbf{99.16}  \\ \cline{2-6}
                        & \multirow{4}{*}{Dashboard}        & Val Accuracy  &94.40  &96.44  &\textbf{98.30}  \\  \cline{3-6}
                        &                                   & Precision &98.91  &98.36  &\textbf{99.18}  \\ \cline{3-6}
                        &                                   & Recall    &94.47  &96.99  &\textbf{98.82}  \\ \cline{3-6}
                        &                                   & F1-Score  &96.62  &97.60  &\textbf{99.00}   \\ \cline{1-6}

\multirow{12}{*}{Car 2} & \multirow{4}{*}{Below Suspension} & Val Accuracy  &92.76  &90.78  &\textbf{97.40}  \\ \cline{3-6}
                        &                                   & Precision &97.67  &\textbf{98.12}  &97.50  \\ \cline{3-6}
                        &                                   & Recall    &93.26  &90.43  &\textbf{98.57}  \\ \cline{3-6}
                        &                                   & F1-Score  &95.34  &93.89  &\textbf{98.02}  \\ \cline{2-6}
                        & \multirow{4}{*}{Above Suspension} & Val Accuracy  &91.86  &87.27  &\textbf{96.49}  \\  \cline{3-6}
                        &                                   & Precision &\textbf{98.01}  &94.34  &96.96  \\ \cline{3-6}
                        &                                   & Recall    &91.96  &89.38  &\textbf{98.81}  \\ \cline{3-6}
                        &                                   & F1-Score  &94.86  &91.43  &\textbf{97.88}  \\ \cline{2-6}
                        
                        & \multirow{4}{*}{Dashboard}        & Val Accuracy  &91.24  &85.77  &\textbf{96.75}  \\  \cline{3-6}
                        &                                   & Precision &\textbf{99.14}  &94.39  &98.22  \\ \cline{3-6}
                        &                                   & Recall    &89.82  &87.71  &\textbf{97.81}  \\ \cline{3-6}
                        &                                   & F1-Score  &94.18  &90.63  &\textbf{98.01}  \\  \cline{1-6}
\multirow{12}{*}{Car 3} & \multirow{4}{*}{Below Suspension} & Val Accuracy  &94.76   &\textbf{98.24}  &97.08  \\  \cline{3-6}
                        &                                   & Precision &\textbf{98.72}  &98.07  &97.73  \\ \cline{3-6}
                        &                                   & Recall    &95.30  &\textbf{99.89}  &98.52  \\ \cline{3-6}
                        &                                   & F1-Score  &96.82  &\textbf{98.97}  &98.09  \\  \cline{2-6}
                        & \multirow{4}{*}{Above Suspension} & Val Accuracy  &95.20  &94.58  &\textbf{99.39}  \\  \cline{3-6}
                        &                                   & Precision &97.88  &97.28  &\textbf{99.78}  \\ \cline{3-6}
                        &                                   & Recall    &96.48  &96.40  &\textbf{99.43}  \\ \cline{3-6}
                        &                                   & F1-Score  &97.11  &96.83  &\textbf{99.60}  \\  \cline{2-6}
                        & \multirow{4}{*}{Dashboard}        & Val Accuracy &96.49  &97.35  &\textbf{98.86 } \\   \cline{3-6}
                        &                                   & Precision &98.74  &97.00  &\textbf{99.56}  \\ \cline{3-6}
                        &                                   & Recall    &97.06  &\textbf{100}  &98.99  \\ \cline{3-6}
                        &                                   & F1-Score  &97.86  &87.70  &\textbf{99.27}  \\ \hline
\end{longtable}
}

\begin{table}[H]
\centering
\scriptsize
\caption{Average Metrics for Cars 1, 2, and 3 (3-BiLSTM, 3-TCN, ETLNet)}
\label{tab:3}
\begin{tabular}{|c|c|c|c|c|}
\hline
\textbf{Car} & \textbf{Metric} & \textbf{3-BiLSTM} & \textbf{3-TCN} & \textbf{ETLNet} \\ \hline
\multirow{4}{*}{Car 1} & Val Accuracy  &94.71  &94.07  &\textbf{98.56}  \\  \cline{2-5}
                       & Precision &98.51  &97.94  &\textbf{98.91}  \\ \cline{2-5}
                       & Recall    &95.26  &94.28  &\textbf{99.07}  \\ \cline{2-5}
                       & F1-Score  &96.82  &95.91  &\textbf{98.99}  \\  \cline{1-5}
\multirow{4}{*}{Car 2} & Val Accuracy  &91.95  &87.94  &\textbf{96.88}  \\ \cline{2-5}
                       & Precision &\textbf{98.27}  &95.62  &97.56  \\  \cline{2-5}
                       & Recall    &91.68  &89.17  &\textbf{98.40}  \\  \cline{2-5}
                       & F1-Score  &94.79  &91.98  &\textbf{97.97}  \\ \cline{2-5}
\multirow{4}{*}{Car 3} & Val Accuracy  &95.48  &96.72  &\textbf{98.44}  \\ \cline{1-5}
                       & Precision &98.45  &97.45  &\textbf{98.76}  \\ \cline{2-5}
                       & Recall    &96.28  &98.76  &\textbf{98.98}  \\ \cline{2-5}
                       & F1-Score  &97.26  &94.50  &\textbf{98.99}  \\ \hline
\end{tabular}
\end{table}

\vspace{-25pt}
\begin{figure}[H]
    \centering
    \includegraphics[width=1\linewidth]{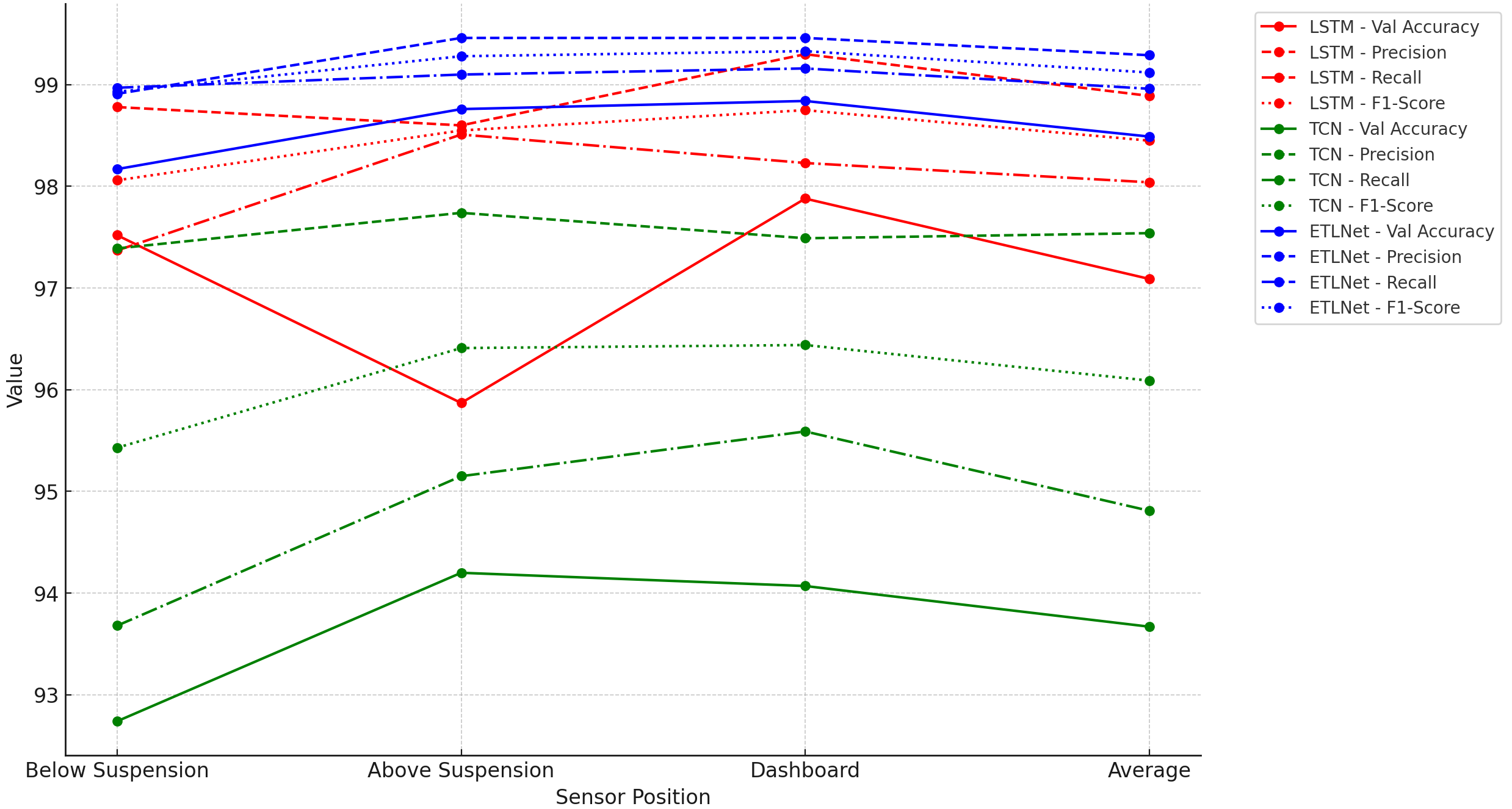}
    \caption{Model Performance}
    \label{fig:enter-label}
\end{figure}

\vspace{-30pt}

\begin{table}[H]
\centering
\small
\caption{Performance Evaluation of Various Models Through Aggregated Data Across All Cars}
\label{tab:4}
\begin{tabular}{|c|c|c|c|c|}
\hline
\textbf{Metric} & \textbf{Sensor Positions} & \textbf{3-BiLSTM} & \textbf{3-TCN} & \textbf{ETLNet} \\ \hline
\multirow{3}{*}{Val Accuracy}  & Below Suspension &97.52  &92.74  &\textbf{98.17}  \\  \cline{2-5}
                               & Above Suspension &95.87  &94.20  &\textbf{98.76}  \\  \cline{2-5}
                               & Dashboard        &97.88  &94.07  &\textbf{98.84}  \\  \cline{2-5}
                               & Average               &97.09  &93.67  &\textbf{98.49}  \\ \cline{1-5}
\multirow{3}{*}{Precision}     & Below Suspension &98.78  &97.39  &\textbf{98.91}  \\  \cline{2-5}
                               & Above Suspension &98.60  &97.74  &\textbf{99.46}  \\  \cline{2-5}
                               & Dashboard        &99.30  &97.49  &\textbf{99.46}  \\  \cline{2-5}
                               & Average               &98.89  &97.54  &\textbf{99.29}  \\ \cline{1-5}
\multirow{3}{*}{Recall}        & Below Suspension &97.37  &93.68  &\textbf{98.97}  \\   \cline{2-5}
                               & Above Suspension &98.51  &95.15  &\textbf{99.10}  \\   \cline{2-5}
                               & Dashboard        &98.23  &95.59  &\textbf{99.16}  \\  \cline{2-5}
                               & Average               &98.04  &94.81  &\textbf{98.96}  \\ \cline{1-5}
\multirow{3}{*}{F1-Score}      & Below Suspension &98.06  &95.43  &\textbf{98.93}  \\  \cline{2-5}
                               & Above Suspension &98.55  &96.41  &\textbf{99.28}  \\  \cline{2-5}
                               & Dashboard        &98.75  &96.44  &\textbf{99.33}  \\ \cline{2-5}
                               & Average               &98.45  &96.09  &\textbf{99.12}   \\ \hline
\end{tabular}
\end{table}

\vspace{-20pt}

\begin{table}[H]
\centering
\small

\caption{Comparative Analysis of Model Performance Across Studies}
\label{tab:5}
\begin{tabular}{|c|c|c|c|c|c|}
\hline
 \textbf{Author} & \textbf{Model} & \textbf{Accuracy} & \textbf{Precision } & \textbf{Recall} & \textbf{F1-Score} \\ 
  \textbf{} & \textbf{} & \textbf{(\%)} & \textbf{(\%)} & \textbf{(\%)} & \textbf{(\%)}\\
 \hline

 Souza, et al. \cite{13}  & DWT, LCSS & 81.13 & 87 & 93 & 90 \\ \hline
 Dey, et al.\cite{4} & SVM, RF, RT & 92 &- &- & 92\\ \hline

Kumar, et al.\cite{15} &ANN &98  &98 &98 &98   \\ \hline
Kempaiah et al.\cite{16}  &LSTM &80 &81 & - & - \\ \hline
Salman et al.\cite{25}  &CNN &92 &81 & 92 & 92 \\ \hline
Menegazzo et al.\cite{6}  & CNN-LSTM $\leftrightarrow$ & 98.59 & 99.64 & 98.75 & 99.19 \\  \hline
\textbf{Proposed Model} & \textbf{ETLNet $\rightarrow$} & \textbf{98.84} & \textbf{99.46} & \textbf{99.19} & \textbf{99.33} \\

\hline
\end{tabular}
\end{table}

\vspace{-10pt}
\bigskip
\noindent \textbf{Icon Legend:}\\
$\leftrightarrow$: Trained on data from both sides of the car. Total parameters: 1408801\\
$\rightarrow$: Trained only on data from the right side of the car. Total parameters: 645377

\section*{Ablation Study}
\label{sec:5}
\subsection*{Model Performance Comparison}

Table \ref{tab:ab_model_performance} presenting the results of the ablation study across different window sizes:

\vspace{-15pt}

\begin{table}[H]
\centering
\scriptsize
\caption{Validation accuracy and F1-scores for each model across various window sizes, including total parameters, graphically shown in Fig \ref{fig:ab_model_comparisons}}

\begin{tabular}{|l|c|c|c|c|c|c|}
\hline
\textbf{Model Description} & 
\begin{tabular}[c]{@{}c@{}}\textbf{100} \\ Val Acc \\ F1-Score\end{tabular} & 
\begin{tabular}[c]{@{}c@{}}\textbf{200} \\ Val Acc \\ F1-Score\end{tabular} & 
\begin{tabular}[c]{@{}c@{}}\textbf{300} \\ Val Acc \\ F1-Score\end{tabular} & 
\begin{tabular}[c]{@{}c@{}}\textbf{400} \\ Val Acc \\ F1-Score\end{tabular} & 
\begin{tabular}[c]{@{}c@{}}\textbf{500} \\ Val Acc \\ F1-Score\end{tabular} & 
\begin{tabular}[c]{@{}c@{}}\textbf{Total} \\ \textbf{Params}\end{tabular} \\
\hline

{\textbf{Base Model}} & 
\begin{tabular}[c]{@{}c@{}}0.93928 \\ 0.94901\end{tabular} & 
\begin{tabular}[c]{@{}c@{}}0.95957 \\ 0.97285\end{tabular} & 
\begin{tabular}[c]{@{}c@{}}0.98100 \\ 0.98849\end{tabular} & 
\begin{tabular}[c]{@{}c@{}}0.98849 \\ 0.9933\end{tabular} & 
\begin{tabular}[c]{@{}c@{}}0.97314 \\ 0.98461\end{tabular} & 
645377 \\
\hline

Single TCN Layer Model & 
\begin{tabular}[c]{@{}c@{}}\textbf{0.94130} \\ 0.95053\end{tabular} & 
\begin{tabular}[c]{@{}c@{}}0.93869 \\ 0.95866\end{tabular} & 
\begin{tabular}[c]{@{}c@{}}0.92576 \\ 0.95281\end{tabular} & 
\begin{tabular}[c]{@{}c@{}}0.91333 \\ 0.94572\end{tabular} & 
\begin{tabular}[c]{@{}c@{}}0.88843 \\ 0.93269\end{tabular} & 
100993 \\
\hline

Dual TCN Layer Model & 
\begin{tabular}[c]{@{}c@{}}0.93155 \\ 0.94161\end{tabular} & 
\begin{tabular}[c]{@{}c@{}}0.94704 \\ 0.96435\end{tabular} & 
\begin{tabular}[c]{@{}c@{}}0.92701 \\ 0.95367\end{tabular} & 
\begin{tabular}[c]{@{}c@{}}0.92563 \\ 0.95492\end{tabular} & 
\begin{tabular}[c]{@{}c@{}}0.89556 \\ 0.93744\end{tabular} & 
364801 \\
\hline

Reduced Feature Model & 
\begin{tabular}[c]{@{}c@{}}0.93199 \\ 0.94193\end{tabular} & 
\begin{tabular}[c]{@{}c@{}}0.95599 \\ 0.97072\end{tabular} & 
\begin{tabular}[c]{@{}c@{}}0.97218 \\ 0.98307\end{tabular} & 
\begin{tabular}[c]{@{}c@{}}0.97910 \\ 0.98784\end{tabular} & 
\begin{tabular}[c]{@{}c@{}}0.97050 \\ 0.98306\end{tabular} & 
644801 \\
\hline

LSTM Replacement Model & 
\begin{tabular}[c]{@{}c@{}}0.94077 \\ \textbf{0.95001}\end{tabular} & 
\begin{tabular}[c]{@{}c@{}}0.95937 \\ 0.97292\end{tabular} & 
\begin{tabular}[c]{@{}c@{}}0.98159 \\ 0.98878\end{tabular} & 
\begin{tabular}[c]{@{}c@{}}0.97759 \\ 0.98688\end{tabular} & 
\begin{tabular}[c]{@{}c@{}}0.97655 \\ 0.98681\end{tabular} & 
694785 \\
\hline

Triple TCN BiLSTM Model & 
\begin{tabular}[c]{@{}c@{}}0.93448 \\ 0.94492\end{tabular} & 
\begin{tabular}[c]{@{}c@{}}\textbf{0.96263} \\ \textbf{0.97508}\end{tabular} & 
\begin{tabular}[c]{@{}c@{}}\textbf{0.98380} \\ \textbf{0.99012}\end{tabular} & 
\begin{tabular}[c]{@{}c@{}}\textbf{0.99216} \\ \textbf{0.99544}\end{tabular} & 
\begin{tabular}[c]{@{}c@{}}\textbf{0.98796} \\ \textbf{0.99313}\end{tabular} & 
843265 \\
\hline
\end{tabular}

\label{tab:ab_model_performance}
\end{table}

\begin{figure}[h]
\centering
\includegraphics[width=01\textwidth]{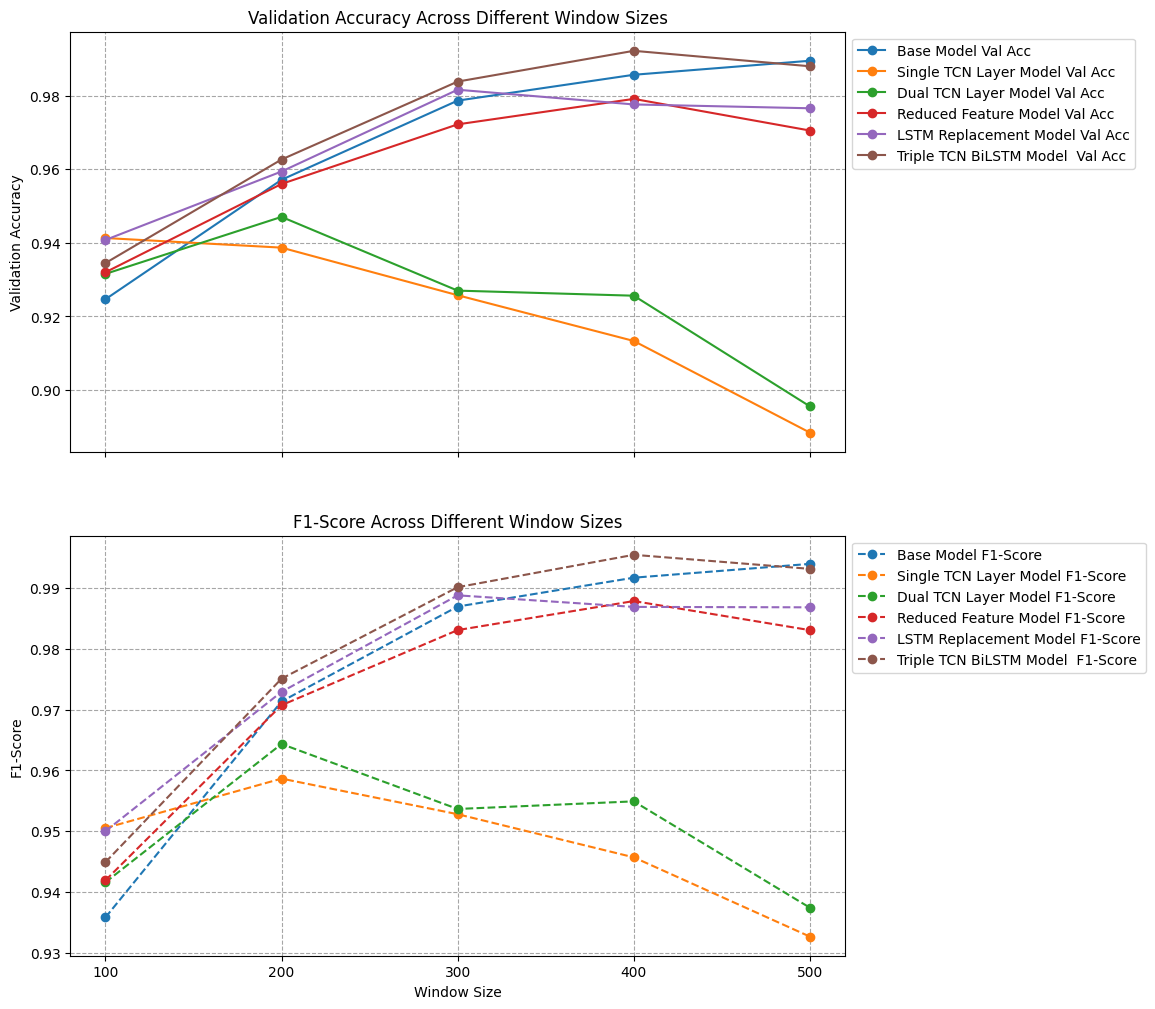}
\caption{Line graphs showing the validation accuracy and F1-scores across different window sizes for each model.}
\label{fig:ab_model_comparisons}
\end{figure}
\vspace{-15pt}

\subsection*{Base Model Proficiency}
The Base Model, equipped with a full suite of features and multiple TCN layers, consistently demonstrated high validation accuracy and F1-scores across all window sizes. This robust performance highlights its adeptness at capturing and processing complex temporal patterns, making it a reliable benchmark for assessing other models.

\subsection*{Impact of Simplification}
\begin{itemize}
    \item \textbf{Single TCN Layer Model:} Showed that reducing the model to one TCN layer significantly hampered its ability to maintain high performance, especially in larger window sizes. This dip suggests that while simplification reduces computational load, it compromises the model's depth needed for analyzing intricate sequences.
    \item \textbf{Dual TCN Layer Model:} Although slightly more complex than the single-layer variant, still fell short of the base model’s metrics, indicating that even moderate simplifications could lead to underperformance.
\end{itemize}

\subsection*{Feature Reduction Consequences}
The Reduced Feature Model highlighted the importance of each input feature. By excluding gyroscope data, we observed a decline in accuracy and F1-scores at larger windows, reinforcing the value of a comprehensive feature set for maintaining predictive precision.

\subsection*{Alternative Architectural Adjustments}
\begin{itemize}
    \item \textbf{LSTM Replacement Model:} Swapped the bidirectional LSTM for a standard LSTM and added a TCN layer, akin to the base model. This configuration offered mixed results, with some improvements in smaller window sizes but generally lower performance in larger ones. This suggests that while alternative architectures can match or slightly improve upon the base in limited contexts, they may not universally provide better outcomes.
    \item \textbf{Triple TCN BiLSTM Model:} Added an extra TCN layer in base model, which led to improvements in certain metrics. The total number of parameters also get increased but with little increase in metrics.  This indicates that additional complexity must be justified by proportional gains in performance.
\end{itemize}

\textbf{Conclusion:} This ablation study effectively highlights the base model’s superiority, not merely in its performance metrics but also in its capacity to consistently handle different data complexities. The exploration through these model variations underscores a critical lesson in machine learning: enhancements or simplifications must be carefully balanced against the model's intended application and performance expectations. Each modification revealed valuable insights into the delicate interplay between model architecture and feature utilization, reaffirming the base model’s design as a solid foundation for reliable and effective predictions in complex temporal data analysis.

\section{Discussion and Conclusion}
\label{sec:6}
In this research paper, the performances of the two existing models along with ablation study, BiLSTM and TCN, have been analyzed, and a new approach, ETLNet, was proposed in the specific context of detecting speed bumps using smartphone sensors. We analyzed vital metrics such as validation accuracy, precision, recall, and F1-score. One of the most revealing aspects of our investigation was how the choice of window size significantly influenced the effectiveness of each model.

In our first experiment, we noted that each model responded differently to changes in window size, impacting their accuracy in speed bump detection. For the BiLSTM model, a window size of 300 consistently produces optimal results across different vehicles and sensor positions. The TCN model, however, presented a more complex story. Its optimal window size varied with each vehicle, suggesting a nuanced sensitivity to the specific characteristics of the data. In contrast, our ETLNet model demonstrated remarkable versatility and strength, maintaining high performance across the board with a uniform window size of 300.

As we extended our analysis in Experiment 2, aggregating sensor position data, the ETLNet model continued to shine, achieving the highest average scores across all vehicles. This spoke volumes about its reliability and ability to deliver consistent results in varied scenarios.

Finally, in Experiment 3, we delved into how well these models generalized their learning across a combined dataset from all vehicles. The results were telling. While the BiLSTM and TCN models performed best with a window size of 300 for specific metrics and sensor positions, the ETLNet model, with a window size of 400, consistently performed better in all respects. This underscored its capability to adapt and excel, even in a diverse and challenging data environment. From the outcomes of our study, it's evident that the dashboard position is particularly effective for this task. In other words, relying solely on data collected from the dashboard also yields promising results.

The system is designed to function through a mobile app, where detected speed bumps will be stored with precise location data, either locally or in the cloud. To register a speed bump for the first time, a vehicle must pass over it. Once recorded, the information will be securely shared with other app users, ensuring that drivers approaching the location receive timely alerts. Each additional pass over the bump will increase the system’s confidence in its presence, improving accuracy. If multiple vehicles no longer detect it, the bump will be automatically removed from the cloud, keeping the database reliable and up-to-date.

Alerts are intended to be generated based on the vehicle’s speed and distance from the bump, giving drivers enough time to slow down smoothly. Notifications will only trigger for speed bumps along the vehicle’s direct path. If a turn is required, the alert will activate only after the vehicle completes the turn and continues straight, avoiding unnecessary distractions. While these features are not yet implemented, the system aims to provide a dynamic, adaptive, and reliable tool that enhances driving safety through timely and accurate alerts.

\subsection{Conclusion}

In this research paper, we proposed a new approach for speed bump detection by carefully studying the existing BiLSTM and TCN architectures and proposing a new approach - ETLNet. This study has illuminated the pivotal role of window size in model performance for speed bump detection and highlighted the ETLNet model's standout capabilities. We've demonstrated the importance of fine-tuning model parameters to suit specific data conditions through careful experimentation and analysis. This principle could be invaluable in enhancing sensor-based vehicular systems. Our findings offer a clear pathway for future research and development in this dynamic field.

\subsubsection{Acknowledgements} 
I thank Professor Aldo von Wangenheim of the Federal University of Santa Catarina for generously sharing his esteemed research. Additionally, I am grateful to the Computer Vision and Biometrics Lab (CVBL) at IIIT - Allahabad for providing access to their GPU resources. The availability of these sophisticated computational tools was crucial during the experimental stage of my research.


%
%
%
%

\end{document}